# Prediction of single well production rate in water-flooding oil fields driven by the fusion of static, temporal and spatial information

Chao MIN, Yijia WANG, Huohai YANG and Wei ZHAO

*Abstract*—It is very difficult to forecast the production rate of oil wells as the output of a single well is sensitive to various uncertain factors, which implicitly or explicitly show the influence of the static, temporal and spatial properties on the oil well production. In this study, a novel machine learning model is constructed to fuse the static geological information, dynamic well production history, and spatial information of the adjacent water injection wells. There are 3 basic modules in this stacking model, which are regarded as the encoders to extract the features from different types of data. One is Multi-Layer Perceptron, which is to analyze the static geological properties of the reservoir that might influence the well production rate. The other two are both LSTMs, which have the input in the form of two matrices rather than vectors, standing for the temporal and the spatial information of the target well. The difference of the two modules is that in the spatial information processing module we take into consideration the time delay of water flooding response, from the injection well to the target well. In addition, we use Symbolic Transfer Entropy to prove the superiorities of the stacking model from the perspective of Causality Discovery. It is proved theoretically and practically that the presented model can make full use of the model structure to integrate the characteristics of the data and the experts' knowledge into the process of machine learning, greatly improving the accuracy and generalization ability of prediction.

## I. INTRODUCTION

To maintain the production of old waterflooding oilfields, it is necessary to design a reasonable development and adjustment plan, the foundation of which is the accurate forecasting of the well production rates. However, due to the heterogeneity of the reservoirs and the complexity of old oilfields, it is very difficult to grasp the development law of a single oil well.

Comparing with the single well, there are many efficient methods to predict the output of a whole oilfield or block, as the uncertainties for an oilfield can be eliminated by the addition of the dada of single wells. These methods, such as empirical formulas, hydrodynamic formulas, material balance equations, and the commonly used time series analysis methods, can also be employed to forecast the single well output, but the performance is always not satisfying since the production of a single well is sensitive to various uncertain factors.

The most commonly used methods to predict an oil well production rate is based on the trend lines of the production history or time series methods. The reason is that the internal and the external influences to a single well production can be implicitly reflected in the dynamic variation of its production. However, these methods are too simplified to handle the complex situation of oil production operation, and are not applicable in the adjustment of oilfield development plans. The mechanism based reservoir simulation method is accurate enough but relies too much on geological modeling and history matching, which is time consuming and professional manpower intensive. The traditional data-driven methods based on machine learning can overcome the aforementioned two defects, but have poor generalization performance to forecast the dynamic production of an oil well. The reason is that sometimes we just input the data all we can get into the machine learning models indiscriminately by some routinely data mining procedure. In these methods the mutual independence of the input data samples is a tacitly acquiescence, but not always true in practical.

Firstly, the production of a single well is related with its underground geological and reservoir properties. But in many cases we can only obtain some limited information of the target well or the average values of the wells in the whole oilfield block. Thus, the result is often unsatisfying because of the heterogeneity of the reservoir and the great differences between the oil wells.

Secondly, the dynamic influence factors to the well production rate are highly time related with each other. If we ignore this temporal dependence of the data, we will always obtain a model with good fitting performance but bad prediction ability.

In addition, in a water flooding oilfield the production of a single well is also influenced by the water injection scheme of adjacent water injection wells. The key is to make it clear of the injection-production response relationship between the target well and adjacent water injection wells, which has a delay of a few months in the data.

In this study, we construct a novel stacking model to fuse the above multi-source information based on the ideology of model agnostic. The experts' experience of dynamic analysis is integrated into the process of model construction and learning through the design of the model structure, which greatly improves the fitting and generalization ability. Each module of the model is not regarded as a purely fitting function, but as a feature extraction tool for three different kinds of information. The fully connected layer is employed to fuse the three different characteristics to achieve accurate prediction of single well production finally. From the experimental verification based on real data, we can see that the method proposed in this study can not only predict single well production accurately but also carry on pre-warning to the sudden changes in production, which will be useful for real-time adjustment of development plan.

The rest of this paper is organized as follows. Section II introduces the related research work of single well production prediction in recent years. Section III presents data preprocessing process, including data analysis, factor screening, and procedure of constructing data samples of fused spatial information. The structure of the stacking model and the theoretical basis of its construction are presented in Section IV. Section V discusses the results of a series of experiments using the stacking model. Finally, Section VI is the conclusion of this study.

## II. RELATED WORKS

In recent years, the research of forecasting single oil well production rate is mainly stressed on three kinds of methods: production decline curves, numerical simulation based on mechanism models and machine learning driven by data.

Among them the method based on decline curves depends on the trend of the production rate, which is relatively accurate in short-term prediction. It ignores the influence of the internal and the external factors, which is not applicable in design and adjustment of the development scheme of the target oil wells. Pingyou Li et al.[1] proposed a new production decline model combined with the development characteristics of low-permeability reservoirs. Three decline parameters are employed to represent three decline stages in this new model: the early fully transition period, the medium-term stable transition period and the late-life period. A descending index parameter is also included in the model. A novel 3-segment decline curve analysis (DCA) method was proposed by Murat Fatih Tugan et al.[2], which uses the unique flow regime sequence of shale wells to delineate the period of the three segments used in DCA. The method is based on the widely-used Arps equation and obeys the dominant flow patterns. Ruud Weijermars and Kiran Nandlal[3] proposed a practical tool for production forecasting based on an analytical flow-cell model for multi-stage fractured shale wells using a two-stage Decline Curve Analysis (DCA) method. The flow-cell model can predict production and estimate ultimate recovery (EUR) of newly planned wells, accounting for changes in completion design, total well length, and well spacing. K Jongkittinarukorn et al.[4] used the Arps hyperbolic decrement method to model the decline rate for each layer in multi-layer wells. The novelty of the method is the ability to provide decreasing parameters for each layer at the early stage of production, which is then used for long-term production forecasting, solving the inconsistency problem in history matching.

The numerical simulation method based on mechanism models is of great significance to guide development scheme, which is adopted in most oilfields at present. Y.G. Qu et al.[5] founded out the factors influencing the water content of single well based on geological characters of complicated fault-block reservoirs. By applying a numerical simulation method, an appropriate simulation model of flooding unit was established. Water injection huff-puff was proved to be an effective method to improve the formation pressure and oil production. Bo Han et al.[6] studied the influence of fracture network on the performance of water injection huff-puff technique through reservoir numerical simulation. Two kinds of fracture network models of main fracture-connected model and main fracture-disconnected model were established. The results proved that fracture network has influence on the performance of water injection huff-puff. However, the accuracy of the mechanism-based methods depends on geological model and history matching, which is time-consuming. In view of the current situation that most oilfields in China have to adjust the development plan on a large scale, this kind method cannot meet the needs of rapid design of development scheme.

In recent years, the machine learning methods are widely used in data-based prediction in oil and gas industry. Several researchers[7][8] made improvements based on BP neural network from the aspects of learning speed and convergence degree to predict the single well production rate. Zhang et al.[9] proposed a method of coal-bed methane production prediction based on BP neural network. Starting from the average daily production of coalbed methane single well, they used the method of grey correlation degree to get the main controlling factors of coalbed methane production. For the main controlling factors, they used BP neural network with high fitting accuracy and get a good prediction result. Jun Yang et al.[10] used SVM and SVD to predict the production curve of natural production wells located in the fault zone. Compared with the Arps decline method, the prediction result is not only more efficient than the traditional decline analysis method, but also has certain reference value for the block to formulate a reasonable production system.

Yuan Zhang et al.[11] applied a new machine learning method based on locality preserving projection (LPP) to predict production of low permeability reservoir. The model preserves the local geometric information inherent in petroleum data and can capture its inherent nonlinear characteristics. Zhang Rui and Jia Hu[12] proposed a method based on multivariate time series (MTS) and vector autoregressive (VAR) machine learning models to predict oil well production of water flooding reservoir. The method first uses MTS analysis to optimize injection and production data based on well pattern analysis. Chong Cao et al.[13] proposed a data-driven modeling approach for small sample multivariate oilfield data. The strengths and limitations of widely used data-driven models and their combined models were analyzed in detail. The aforementioned models are based on data independence assumption, ignoring the change of production dynamic information in time, which will result in good fitting ability but poor generalization performance.

Then a class of methods based Recurrent Neural Network(RNN) are employed to deal with the temporal information. Dongyan Fan et al.[14] established a new hybrid model that considers the advantages of linearity and nonlinearity, as well as the influence of manual operations. This integrates an Autoregressive Integrated Moving Average (ARIMA) model and a Long-Short-Term Memory (LSTM) model. The ARIMA model filters linear trends in the production time series data and passes the residual values to the LSTM model. Xuechen Li et al.[15] proposed a new framework using Bidirectional Gated Recurrent Unit (Bi-GRU) and Sparrow Search Algorithm (SSA) to improve the accuracy of the production rate of oil wells. The observations showed that the method outperforms traditional descent curve analysis, time series methods, and one-way recurrent neural networks in terms of accuracy and robustness. N. Chithra Chakra et al.[16] used a higher-order neural network (HONN) to predict oil production, overcoming the limitations of the traditional neural networks by simultaneously representing the linear and nonlinear dependencies of the neural input variables. On this basis, Joko Prasetyo et al.[17] focused on the impact of normalization on HONN models. A non-linear transformation called quantile transformation was proposed to improve prediction accuracy, and the results obtained by various normalization methods were evaluated. Xuechen Li et al.[18] proposed a Gated Recurrent Unit (GRU) -based neural network model to achieve batch production prediction of reservoirs, overcoming the limitations of traditional decline curve analysis and conventional time series forecasting methods. Mayor Pal[19] focused on applying advanced data analytics and deep machine learning methods to time series prediction of injection and production data from subsurface hydrocarbon recovery processes. He first tested the RNN-LSTM algorithm on field data including multiple injection-production well patterns, discussing the complexities in detail associated with data acquisition, analysis, and processing. Masanori Kurihara et al.[20] investigated several models, such as auto-regressive (AR), multilayer perceptron (MLP), and long short-term memory (LSTM) networks. These models are based on static and dynamic parameters and daily fluid production while considering the inverse distance from adjacent wells. Joko Nugroho Prasetyo et al.[21] proposed a machine learning system based on Hybrid Empirical Mode Decomposition-Backpropagation High-Order Neural Network (EMD-BP-HONN) for forecasting oil production flowrate. It solves the unique challenges brought by the stationary and non-stationary characteristics of single-well data. In practice we found that the purely RNN methods regard the static geological information as constant dynamic signals, which will cover up the changes of dynamic features, resulting in poor prediction accuracy.

In fact, the production of single well is comprehensively influenced by three aspects of factors: static geological information, dynamic well production history, and spatial information of the adjacent water injection wells. Therefore, it is necessary to consider changing the construction of the model in structure, enhancing the information fusion ability of the model to improve the final fitting and generalization ability. It is found that the stacking neural network model proposed in this study has good performance in fusing multi-source and multi-types of information to forecast single well production, which is rarely discussed in the literatures. DEQI CHEN et al.[22] proposed a stacked Bidirectional Gated Recurrent Unit neural network model. By establishing a multiscale-grid model, a set of key traffic parameters of different scales can be deduced in less time to predict the traffic speed of various-scale road. Anu Jose and Vidya V[23] constructed a stacked Long short-term memory (LSTM) model, which has a certain memory capacity and can accurately analyze sequential data. The stacked LSTM hidden layer enhances the strength of the model and improves the prediction accuracy. Yusera Farooq Khan et al.[24] used Convolutional Neural Network (CNN) to design a hybrid model of CNN & Bidirectional Long-Short Term Memory (Bidirectional LSTM), and proposed a Stacked Deep Dense Neural Network (SDDNN) model for text classification and prediction of Alzheimer's disease. These existing stacking methods are applied in non-petroleum fields and not using the parallel structure to stack the networks as described in this study.

### III. DATA PREPROCESSING

In order to guarantee the performance of the predictive model, we must ensure the data quantity and quality. In practice, due to improper behavior in the recording or operation process, the data obtained should be preprocessed first, including data cleaning, data segmentation, normalization, etc. The following table describes the basic information of the datasets used in this study.

TABLE I
CATEGORIZATION AND DESCRIPTION OF DATA

**STATIC DATA OF OILFIELD BLOCK**

| Parameter | Description | Data Type |
|---|---|---|
| Horizon | A specific position in a stratigraphic sequence | Nominal |
| Top Depth of Perforated Well Section | Bottom interface depth and top interface depth of each layer in perforated interval | Numeric |
| Bottom depth of perforated well section | | Numeric |
| Top depth of shooting oil (gas) layer | Distance from the uppermost (lowermost) part of perforated interval to main casing | Numeric |
| Bottom depth of shooting oil (gas) layer | | Numeric |
| Interpreted thickness | The formation thickness of logging interpretation, generally refers to the sand thickness of oil layer | Numeric |
| Effective thickness | The thickness of oil layer that can produce oil of industrial value under the existing economic and technical conditions | Numeric |
| Converted central depth | Depth from oil and water wellhead to the middle of current production layer | Numeric |
| Original formation pressure | The middle pressure of oil layer is measured in exploration wells when the oil field has not been put into development, | Numeric |
| Original saturation pressure | Pressure at which natural gas begins to separate from crude oil | Numeric |
| Original formation temperature | Measured reservoir temperature under original reservoir conditions | Numeric |
| Geological reserves | The total amount of crude oil or natural gas in the reservoir with the ability to produce oil and gas under the original conditions of the formation | Numeric |

**DYNAMIC DATA OF DEVELOPMENT**

| Parameter | Description | Data Type |
|---|---|---|
| Production days | | Numeric |
| pump deep | | Numeric |
| Pump diameter | Inner diameter of oil well pump barrel. Under the condition of constant stroke and stroke times, increasing the pump diameter can improve the output of pumping wells | Numeric |
| Pump efficiency | The ratio of the actual amount of liquid pumped out by the oil well pump to the theoretical amount of liquid pumped out | Numeric |
| Swept volume | Volume of liquid delivered by the pump in unit time | Numeric |
| Stroke | When the pump diameter is fixed, the output of a pumping well mainly depends on the length of the stroke and the number of strokes | Numeric |
| frequency of stroke | The number of up and down movements of the oil pump piston in the working cylinder per minute | Numeric |
| Oil pressure | Residual pressure of oil well flowing from bottom to wellhead during normal production | Numeric |
| Casing pressure | Wellhead pressure of oil jacket annulus | Numeric |
| Flow pressure | The pressure of the fluid in the middle of the reservoir measured during normal production of the oil well | Numeric |
| Static pressure | After the oil well is put into production, the pressure in the middle of the oil layer measured when the bottom hole pressure returns to stability by short-term shut in | Numeric |
| Dynamic liquid level | The liquid level height in annular space between the tubing and casing during the production in a non-flowing well. According to liquid level height and relative liquid density, the flow pressure of the oil well can be calculated | Numeric |
| Cumulative oil / water production | | Numeric |

TABLE 1
SPATIAL DATA REFLECTING WELL PATTERN CONDITIONS

| Parameter | Description | Data Type |
|---|---|---|
| Horizontal/vertical coordinates | The rectangular coordinate system established with a specified fixed point as the origin | Numeric |
| Monthly water injection | Total monthly water injection volume, including the sum of monthly sewage injection volume and monthly clean water injection volume | Numeric |

OUTPUT

| Parameter | Description | Data Type |
|---|---|---|
| Monthly oil production | | Numeric |

*A. Data categorizing*

The key of a predictive model is to find the potential input-output relationship from the data, while there are many factors that might influence the output of a single oil well in practical production process. Thus in this study we first divide these influencing factors into three categories: static geological information, dynamic oil well production history and spatial information of adjacent water injection wells, so that the input of the presented model can be consistent with the physical cognition to some extent.

As shown in Table 1, the static geological information is the properties of the target oilfield block, including the original formation pressure, geological reserves, effective thickness etc. which ignore the heterogeneity of the reservoir and are average values or coarse information. The dynamic well production history information includes two categories. One is the factors that influence the stable operation and oil production efficiency, such as pump depth, pump efficiency, displacement, stroke, stroke frequency, casing pressure and dynamic liquid level, which reflect the functional status of oil fields. The other is the production dynamic index of oilfield development such as production days, cumulative oil production and cumulative water production. Spatial information mainly reflects the water injection response relationship between oil wells and surrounding water injection wells. Finally, the output of the presented model is monthly oil production of the target oil well.

*B. Data preprocessing for oil well production prediction*

*1) Processing of Dynamic and Static Data*

It is necessary to use appropriate methods to select the main factors that really influence the oil well production to be the input of the model. In this study, the factors influencing monthly oil production are selected according to the data exploration process shown in Figure 1.

The influence of oilfield external environment or anthropogenic misregistration will lead to outliers in the dataset. Therefore, it is necessary to process the outliers in the data first in order to ensure the rationality and validity of the data, so as not to influence the subsequent model training. Here, the boxplot method is employed to identify outliers first, and then the identified outliers are transformed into missing values for processing. The missing values are filled by different methods. If the missing value belongs to static factors, the KNN method is employed to fill the blank. If the missing value belongs to dynamic sequence, the interpolation method is employed to fill it.

Then the univariate analysis is carried out to observe the statistic distribution characteristics of the data. By multi-factor analysis, such as correlation analysis, factor analysis and cluster analysis, we explore the interaction of influencing factors on the predictor. Finally, the embedded feature selection method is employed to determine the model input. Thus the selected input can not only meet the basic requirements of statistical learning, but also in coincident with the practical experience. Finally, 13 representative and reasonable features are selected as the input of the predictive model, as shown in Table 2.

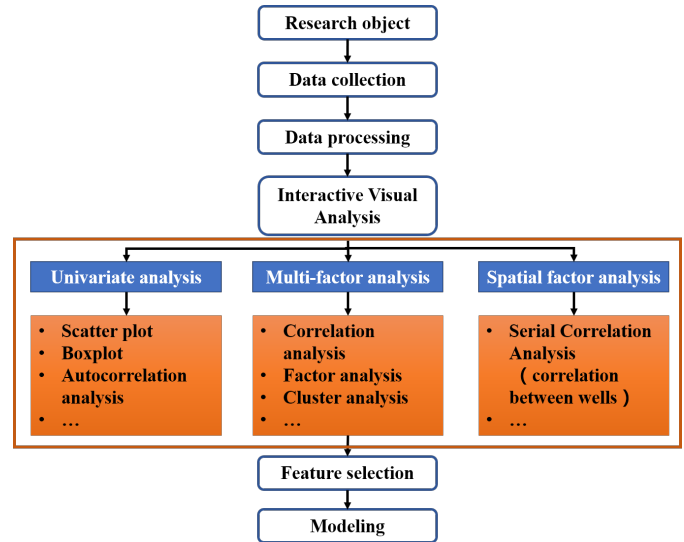

Fig. 1. Exploratory Data Analysis Process

TABLE 2
INPUT VARIABLES OF THE MODEL

| Variable name | Unite |
|---|---|
| Converted central depth | m |
| original formation pressure | MPa |
| original formation temperature | °C |
| geological reserves | t |
| Effective thickness | m |
| Production days | d |
| pump deep | m |
| Pump efficiency | % |
| Swept volume | L |
| stroke | M |
| frequency of stroke | SPM |
| casing pressure | MPa |
| liquid level | m |

*2) Spatial information collection*

Water injection is an efficient technique to enhance oil recovery by injecting water into oil layers to drive the oil and maintain the reservoir pressure. In water flooding oilfields, the water flooding efficiency of water injection wells has significant influence on oil well production. Through the analysis of injection-production effectiveness, we can improve the control degree of the current well pattern on the reservoir, so that the oil well can receive better water injection efficiency and improve the overall development level[25]. However, the water injection is a dynamic process, the response of the production well will have some delay, which makes it difficult to consider the temporal and spatial relationship in the existing machine learning models.

This study uses a serial correlation analysis method to explore the spatial connection between the oil wells and the water wells. Firstly, the spatially adjacent water injection wells of oil production wells are determined according to the geographical coordinates of the well location. Then the cross-correlation coefficient of monthly oil production and monthly water injection of adjacent oil and water wells is calculated. The dynamic input of our model is formulated by the following process considering the temporal and spatial influence of the water injection wells:

(1) Taking the Euclidean distance as a constraint, the water injection wells around the target oil well are obtained according to the geodetic coordinates. Each oil production well corresponds to at least one water injection well;

(2) Select the water injection well with the largest absolute value of the correlation coefficient among the adjacent water injection wells as the corresponding adjacent well of the oil well. The correlation coefficient is obtained by calculating the series cross-correlation coefficient with lag order. The time series of monthly oil production the oil wells and the time series of monthly water injection of the water wells are unified as the following vectors:

$$\begin{cases} Y = [y_{1,t}, y_{2,t}, \cdots, y_{i,t}, \cdots, y_{l_o,t}]^T \\ B = [b_{1,t}, b_{2,t}, \cdots, b_{j,t}, \cdots, b_{l_w,t}]^T \end{cases} (t = 1, 2, \cdots, n) \quad (1)$$

Here $b_{j,t}(j=1,2,\cdots,l_w; t=1,2,\cdots,n)$ are the monthly water injection, $y_{i,t}(i=1,2,\cdots,l_o; t=1,2,\cdots,n)$ are the monthly oil production; When the time interval is $\theta(\theta = t_2 - t_1, t_2 > t_1)$, the covariance[26] between sequences $Y$ and $B$ is:

$$s_{i,j}(\theta) = \frac{1}{n-\theta} \sum_{t=1}^{n-t} (y_{i,t} - \bar{y})(b_{j,t+\theta} - \bar{b}) \quad (2)$$

where $\bar{y} = \frac{1}{n}\sum_{t=1}^{n} y_{i,t}$ is the mean value of the sequence $y_{i,t}(t=1,2,\cdots,n)$ and $\bar{b} = \frac{1}{n}\sum_{t=1}^{n} b_{j,t}$ is the mean value of sequence $b_{j,t}(t=1,2,\cdots,n)$, then the cross-correlation coefficient is:

$$r_{i,j}(\theta) = \frac{s_{i,j}(\theta)}{s^2} = \frac{1}{n-\theta} \sum_{t=1}^{n-\theta} (\frac{y_{j,t} - \bar{y}}{s_{i,j}(\theta)})(\frac{b_{j,t+\theta} - \bar{b}}{s_{i,j}(\theta)}) \quad (3)$$

According to the Pearson linear correlation degree evaluation standard, the monthly water injection data of the water injection well with the highest degree of correlation with the monthly oil production data of each target oil production well is selected. The process is described by the formula as follows:

$$\begin{cases} M_{i,L_w} = \max(|r_{i,1}|, |r_{i,2}|, \cdots, |r_{i,l_w}|) \quad (i=1,2,\cdots,l_o) \\ M_{i,L_w} \geq 0.3 \\ \theta_{i,L_w} < 12 \end{cases} \quad (4)$$

$M_{i,L_w}$ refers to the maximum absolute value of the cross-correlation coefficient with the lag order $\theta_{i,L_w}$, which is calculated from the monthly oil production data of the $i$-th oil production well and the corresponding monthly water injection data of the $L_w$-th water injection well. $\theta_{i,L_w}$ stands for the monthly oil production data of the $i$-th oil production well lags the monthly water injection data of the $L_w$-th water injection well by $\theta$ months. By formula (4), the input/output dataset fusing the spatial information is constructed according to the process shown in Figure 2.

For the oil wells whose adjacent water injection well can be determined by formula (4), the monthly water injection data of the adjacent water injection well are intercepted by moving forward or backward according to the corresponding lag order $\theta_{i,L_w}$. For the oil wells that none of the adjacent water injection wells meet the threshold conditions, inserting the average monthly water injection sequence of its adjacent water wells as the input factor, which is equivalent to giving a background noise.

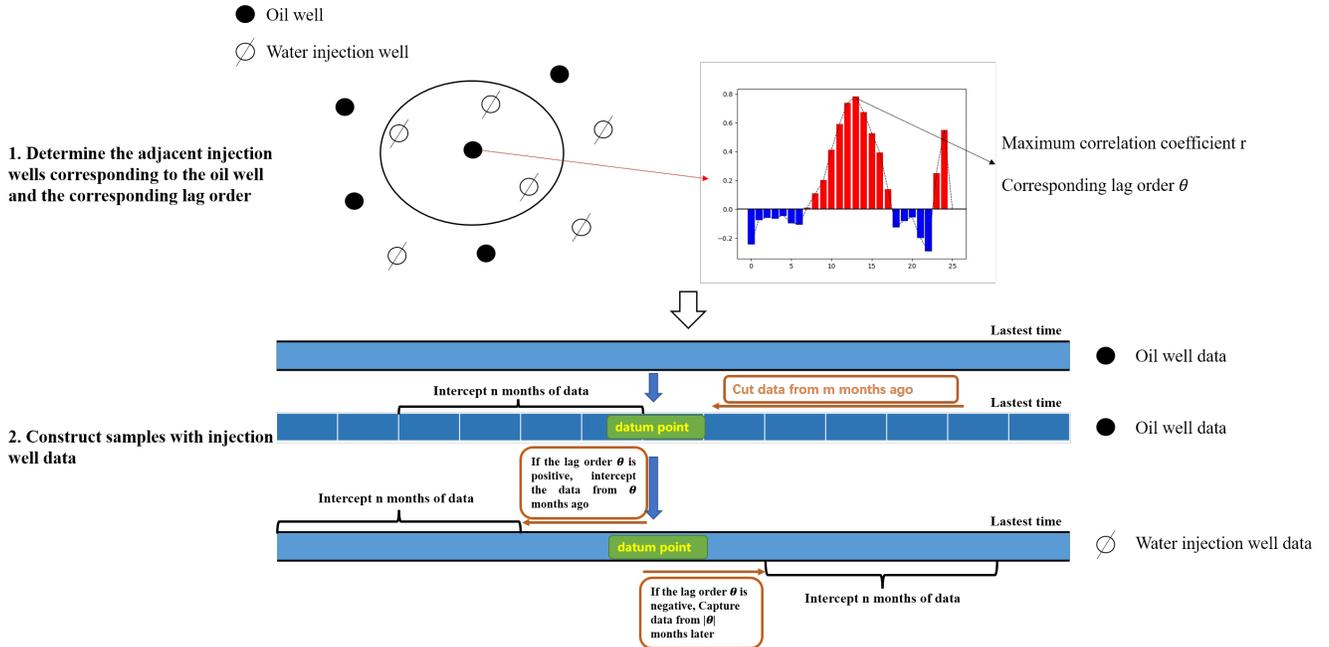

Fig. 2. Exploratory Data Analysis Process

## IV. THEORETICAL BASIS OF THE MODEL

Poor robustness and weak generalization ability are the main defects of the machine learning methods to predict an oil well's production. In addition, the statistic distribution of oilfield data in different development stages and development states are quite different, thus, the mobility of the algorithm is not strong and is not conducive to the accurate prediction of the production in the complex oilfield environment. Based on the idea of meta-learning, this study proposes a predictive model based on BP-LSTM stacking network to predict the oil well production.

Fig.3 shows the process of constructing the stacking model. First, the prediction is performed using the basic machine learning model, where the samples are treated as static and independent. Further, the prediction is performed using the time series model, where the samples are treated as dynamic time series. If the generalization ability of the predictions of both models does not satisfy the threshold, we construct the stacking model to integrate both static and dynamic information. Finally, the water injection information of spatially adjacent injection wells is fused into the sample to bring the model generalization capability to a threshold value. The process takes the machine learning model at each step as input-output simulator, i.e., feature extractor, integrating the update process of data task into the construction process of the stacking model.

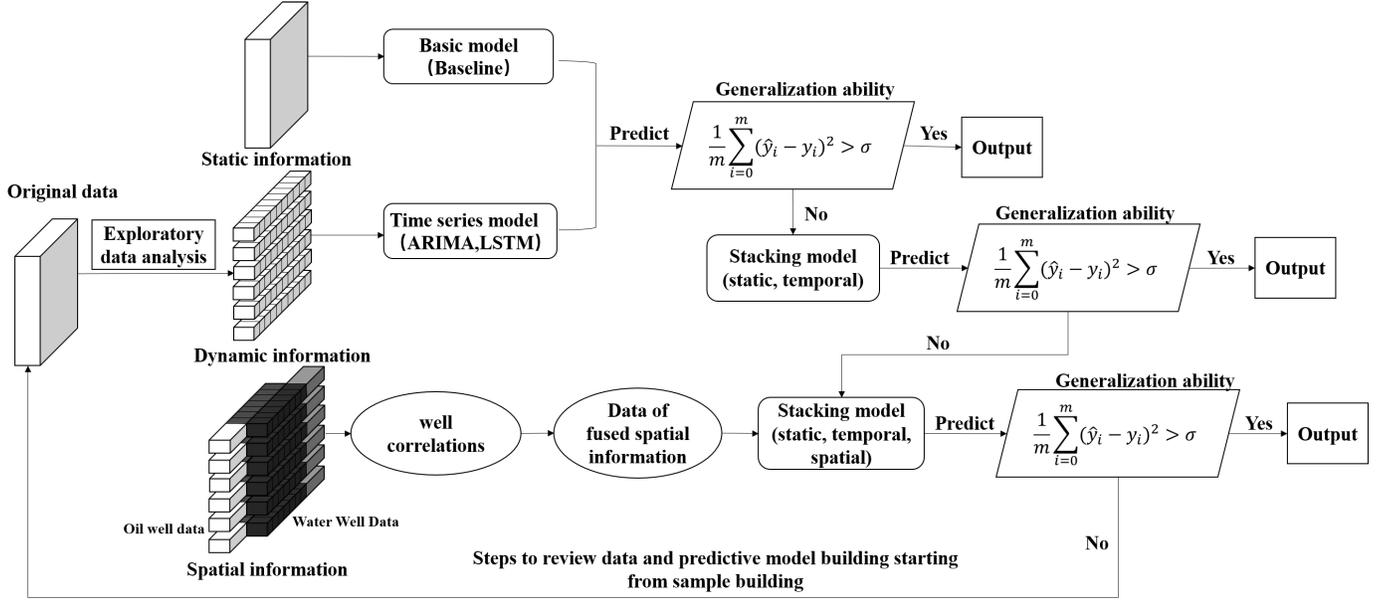

Fig. 3. Flow Chart of the Construction of Oilfield Production Prediction Model

## A. BP-LSTM stacking model

In this study, a BP-LSTM stacking neural network model is proposed with the basic structure shown in Figure 4. The number of neurons in the input layer of BP network is the number of static input features, and the number of neurons in the input layer of LSTM network is the number of dynamic input features. In order to prevent overfitting, a hidden layer is set up, and then the output information of BP network and LSTM network is stacked to the fully connected layer by Add layer. The output layer is the time series data of monthly oil production of an oil well. In this model, multilayer perceptron (MLP), as an encoder, is an effective tool for extracting features from the static information, including underground stratum information and other static factors[27]. Its input is in the form of vector with the components such as geological reserves and effective thickness etc. In addition, LSTM is an important tool for dealing with the time series data, which is applied to extract the variation trends of the dynamic factors as production days and dynamic liquid level etc. Its input is the time series of these factors, and thus in the form of matrix. The input-output mapping relationship of the stacking model is shown in Equation 5.

$$\begin{Bmatrix} X_s \\ X_d \end{Bmatrix} \to \begin{pmatrix} y_1 \\ y_2 \\ \vdots \\ y_T \end{pmatrix} \Rightarrow \begin{Bmatrix} \begin{pmatrix} x_s^1 & x_s^2 & \cdots & x_s^k \end{pmatrix}^\top \\ \begin{pmatrix} x_{d,1}^1 & x_{d,2}^1 & \cdots & x_{d,T}^1 \\ x_{d,1}^2 & x_{d,2}^2 & \cdots & x_{d,T}^2 \\ & & \vdots & \\ x_{d,1}^l & x_{d,2}^l & \cdots & x_{d,T}^l \end{pmatrix} \end{Bmatrix} \to \begin{pmatrix} y_1 \\ y_2 \\ \vdots \\ y_T \end{pmatrix} \quad (5)$$

Where $X_s$ is the input of static information, $X_d$ is the input of dynamic information including temporal and spatial data. $k$ is the number of static features and $l$ is the number of dynamic features. Let $\hat{Y}_T = (\hat{y}_1, \hat{y}_2, \cdots, \hat{y}_T)^\top$ represent the monthly oil production series value predicted by the BP-LSTM stacking model, $Y_T = (y_1, y_2, \cdots, y_T)^\top$ present the true monthly oil production series value, and $T$ represent the length of the input dynamic series. Samples of $m$ wells are used for model training, and samples of $n$ wells are used for model testing. The dimensions of input samples $X_{s,train}$ and $X_{d,train}$ for model training are $m \times 1 \times k$ and $m \times T \times l$ respectively. The dimensions of input samples $X_{s,test}$ and $X_{d,test}$ for model testing are $n \times 1 \times k$ and $n \times T \times l$. Let

$$\hat{Y}_T = f(\hat{X}) \quad (6)$$

$$\hat{X} = \sum_{i=1}^{c} (\hat{X}_s^i + \hat{X}_d^i) * k_i \quad (7)$$

where $X_s^i$ is the output of the $i$-th neuron processed by BP neural network, and $X_d^i$ is the output of the $i$-th neuron processed by LSTM neural network. The number of output channels of the two networks are both $c$. After adding the values of the corresponding neurons of the two networks, the convolution value is calculated with the corresponding convolution kernel $k_i$ for accumulation, and the feature fusion of the two networks is realized. The loss function of the model is as follows:

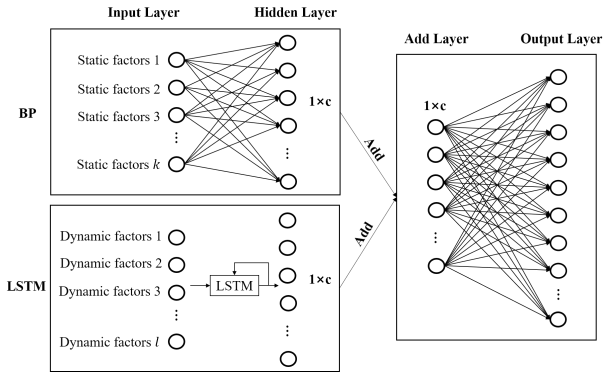

Fig. 4. Structure of BP-LSTM Stacking Network Model

$$\text{loss function} = \frac{1}{n \cdot T} \sum_{j=1}^{n} \sum_{i=1}^{T} (\hat{y}_{i,j} - y_{i,j})^2 \quad (8)$$

## B. Theoretical foundation for the construction of the presented stacking model

Most of the current state-of-art work on machine learning is based on the independent and identically distributed data, which can ensure that the algorithm converges to the lowest achievable risk. If we can get enough data, the machine learning model can perform very well. But in practical problem the data always violates the assumption of independence and identical distribution, thus the performance of machine learning model is usually poor[28]. To fix this problem, in many research the mechanism is employed to combine with the machine learning model. In this study, as the influence of the static and the dynamic factors on single well production have different mechanism, we deal with the static and dynamic information separately by constructing the stacking model. This idea is inspired by the theory of causality discovery. The model of stacking structure, to some extent, reflects the mechanism of the process of the dynamic analysis of oil well production by professionals. Here the symbolic transfer entropy (STE) is employed to illustrate the effect of the stacking structure, which can quantitatively describe the coupling strength and asymmetric driving response relationship of two systems[29].

STE is applied to non-stationary time series and can measure the non-linear causal response relationship between variables. Let two time series as $c_i = c(i)$ and $c_j = c(j)$ $i, j = 1, 2, \cdots, N$, and define symbols by reordering time series $c_i$ and $c_j$. For a given but arbitrary $i$, $m$, the amplitude values $C_i = \{x(i), x(i+l), \cdots, x(i+(m-1)l)\}$ are arranged in ascending order $\{c(i+(k_{i1}-1)l) \leq c(i+(k_{i2}-1)l) \leq \cdots \leq c(i+(k_{im}-1)l)\}$, where $l$ is the time delay and $m$ is the embedding dimension. When the amplitude values are equal, rearrange according to the relevant index $k$, that is, for $c(i+(k_{i1}-1)l) = c(i+(k_{i2}-1)l)$ we write $c(i+(k_{i1}-1)l) \leq c(i+(k_{i2}-1)l)$, if $k_{i1} < k_{i2}$, to ensure that each $C_i$ uniquely maps to $m!$ possible permutations.

A symbol is denoted by $\hat{c}_i \equiv (k_{i1}, k_{i2}, \cdots, k_{im})$, and the relative frequency of the symbol is employed to estimate the joint probability and conditional probability of the permutation index sequence.

**Definition 1 (Symbolic Transfer Entropy):** Given the symbol sequences $\{\hat{c}_i\}$ and $\{\hat{c}_j\}$, the symbol transfer entropy (STE) is defined as:

$$STE_{c_j \to c_i} = \sum p(\hat{c}_{i+\delta}, \hat{c}_i, \hat{c}_j) \log \frac{p(\hat{c}_{i+\delta} | \hat{c}_i, \hat{c}_j)}{p(\hat{c}_{i+\delta} | \hat{c}_i)} \quad (9)$$

where the sum traverses all symbols, and $\delta$ representing the time step. The log is based on 2. $T_{c_i \leftarrow c_j}$ is defined in complete analogy.

The directivity index $STE = (STE_{c_i \to c_j} - STE_{c_j \to c_i})$ quantifies the preferred direction of information flow, which is expected to attain positive values for unidirectional couplings with $c_i$ as the driver and negative values for $c_j$ driving $c_i$. For symmetric bidirectional coupling, we expect $STE = 0$.

Here, the delay coefficient $\delta$ is determined by using the mutual information method, and the embedding dimension $m$ is determined by using the geometric invariant method (correlation dimension). We calculate the STE values between the static data, dynamic data, and the combined static-dynamic data with monthly oil production respectively. The results are shown in Figure 5. The horizontal coordinate is the delay time of monthly oil production compared to the factors. It can be seen from the figure that the STE calculated by the combined data (dynamic and static) has the highest value, comparing with the results by either static or dynamic data alone. This quantity result verified that the stacking model in this study has better performance than the traditional machine learning model, which integrates both the overall condition of the static factor representation and the time-series change state of the dynamic factor representation.

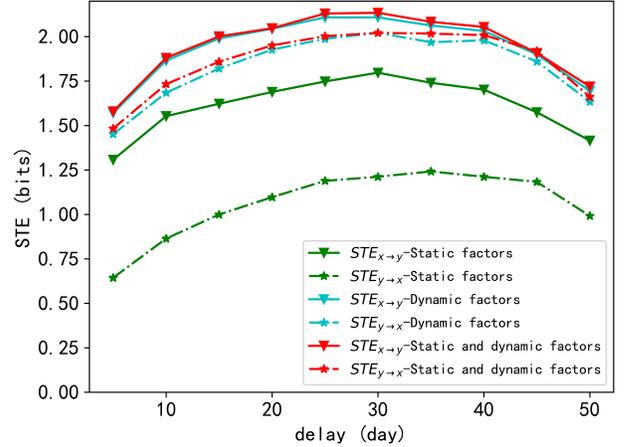

Fig. 5. Calculation Results of Symbolic Transfer Entropy (the Curve with Lower Triangle is the STE Value of Factor $X$ to Production $Y$, and the Curve with Asterisk is the STE Value of Production $Y$ to Factor $X$).

## V. EXPERIMENTS AND RESULTS

### A. Experimental Setup

A block of some water flooding oilfield in China is selected for experiment, which has 124 wells and enough historic development data. The dataset used in this experiment containing 30 oil wells, with 2520 sample instances and 13 features.

In this study, $R^2$ Score, Root Mean Square Error (RMSE) and Mean Absolute Error (MAE) are employed as evaluation indicators to assess the algorithm's prediction performance. The Root Mean Square Error represents the sample standard deviation of the bias between the observed value and the predicted value, indicating the discrete degree of samples. Mean Absolute Error represents the average of the absolute errors between the predicted value and the observed value. RMSE penalizes large errors more than small errors, and outliers or large errors lead to larger RMSE, while MAE does not necessarily increase with the increase of the variance of the margin of error. These evaluation indicators are defined as follows:

$$R^2 score = 1 - \frac{SS_{res}}{SS_{tot}} \quad (10)$$

$$RMSE = \sqrt{\sum_{t=1}^{T} \frac{(\hat{y}_t - y_t)^2}{T}} \quad (11)$$

$$MAE = \frac{1}{T} \sum_{t=1}^{T} |\hat{y}_t - y_t| \quad (12)$$

Where $SS_{res} = \sum_{t=1}^{T} (\hat{y}_t - y_t)^2$, $SS_{tot} = \sum_{t=1}^{T} (\bar{y}_t - y_t)^2$. $\hat{y}_t$ is the predicted value and $y_t$ is the true value at time $t$ over a period of $T$ time steps.

## B. Experimental Results

### 1. Model Comparison

According to the result of autocorrelation analysis of time series data of dynamic features, as shown in Figure 6, the autocorrelation coefficients above 60 orders lagged are beyond the confidence boundary. Therefore, the most recent 60-month data of each well was segmented to construct the sample. For wells with less than 60 months of data, fill in 0 in front of the sequence as no production. Taking the data of one well as a sample, the data of 30 wells constitute the whole sample set.

According to the 13 feature variables obtained by data exploration process, the single well production prediction model is established. Three machine learning algorithms: random forest, LSTM neural network and the BP-LSTM stacking network proposed in this study, are employed to predict. Among them, random forest stands for static predictive model, which treats each sample as independent and unconnected; LSTM neural network stands for dynamic predictive model, which can extract the characteristics of time series changes in data. The prediction performance of these models in several wells is shown in Table 3.

Figure 7 shows the prediction results of the three models in Well 1 and Well 2, respectively. It can be seen from the figure that the predictive deviation of the random forest model is relatively large. The reason is that the forecasting method belongs to static learning, failing to consider the changes and connections in the dynamic trend of a well over a period of time. The prediction result of LSTM network is much better than random forest, indicating that the LSTM network extracts the changing characteristics of dynamic time series factors. However, according to the learning curve of LSTM network shown in Figure 8(a), we can see that the model has overfitting phenomenon and poor generalization ability. The stacking model integrates the influence of static feature vectors and the dynamic time series of change trends on the prediction indicators, and its predictive accuracy is significantly improved compared with the random forest and LSTM. According to the learning curve of the BP-LSTM stacking network shown in Figure 8(b), it is known that the generalization ability of the model is up to the standard.

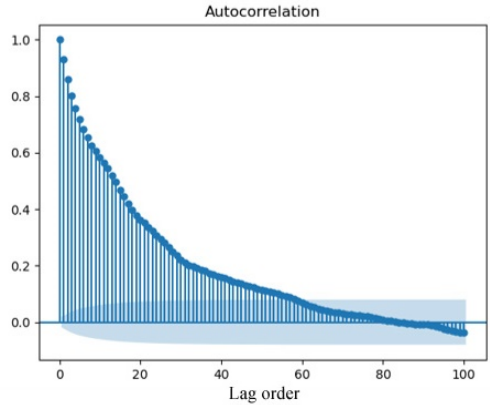

Fig. 6. Autocorrelation Analysis of Monthly Oil Production of the Single Well

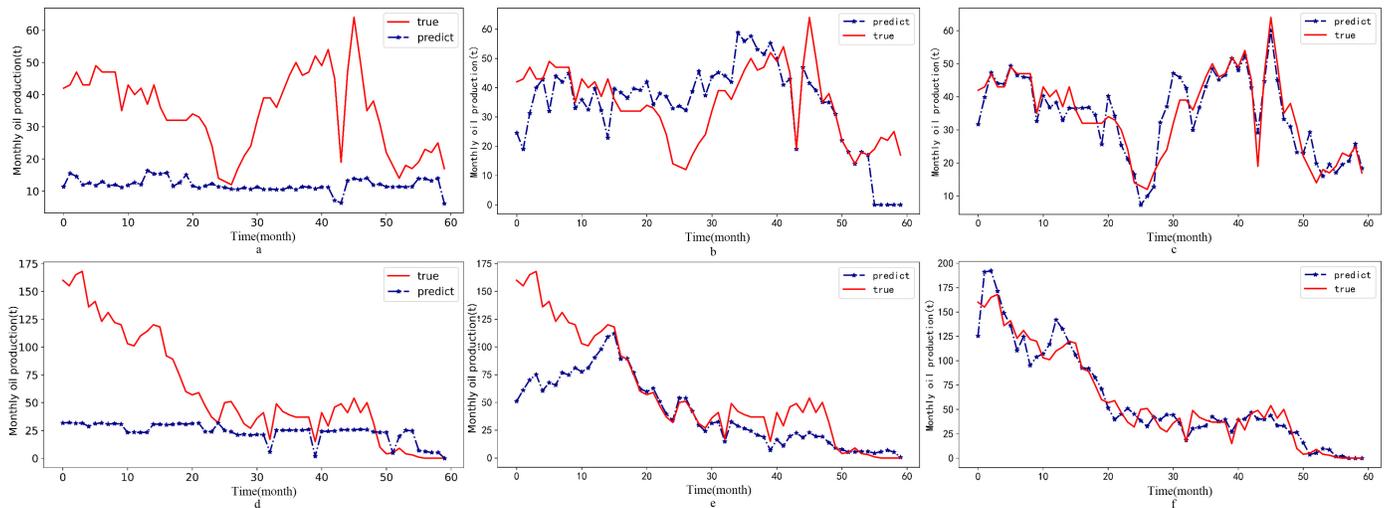

Fig. 7. Three Models' Prediction Results of Two Wells (a, b and c are the Prediction Results of Random Forest, LSTM and BP-LSTM Stacking Network in Well 1, Respectively; d, e and f are the Prediction Results of Random Forest, LSTM Neural Network and BP-LSTM Stacking Network in Well 2, Respectively).

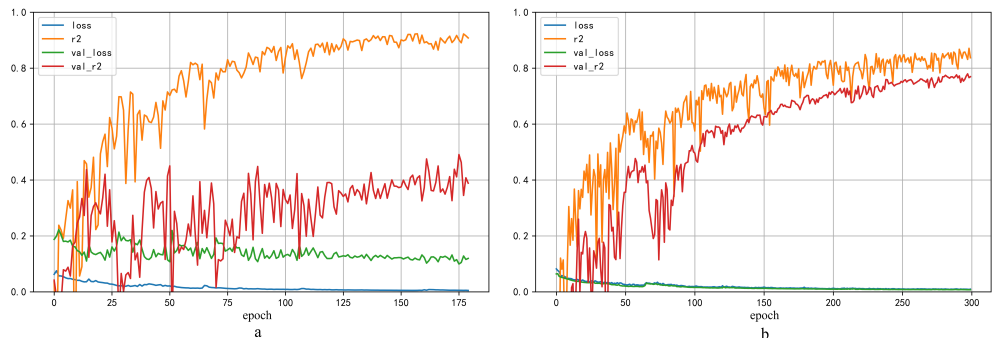

Fig. 8. Learning Curve of LSTM Neural Network(a) and BP-LSTM Stacking Neural Network (b).

TABLE 3
EVALUATION INDICATORS PREDICTED BY THE THREE MODELS IN TWO WELLS

|  | Well 1 | | | Well 2 | | |
| --- | --- | --- | --- | --- | --- | --- |
| Model | R2 | MAE | RMSE | R2 | MAE | RMSE |
| *Random Forest* | -3.39442 | 22.85986 | 25.86033 | -0.33844 | 39.45848 | 55.58873 |
| *LSTM* | 0.485315 | 6.662456 | 8.850213 | 0.505758 | 20.59217 | 33.77983 |
| ***Our Model*** | **0.83159** | **3.468651** | **6.906736** | **0.938746** | **3.381062** | **4.988091** |

2. Discussion on the stacking model of fused data

In this section, we compare the results of the model trained by BP-LSTM stacking neural network to predict the production for samples without and with fusion of water injection information. Following the spatial information collection process described in Section 2, a sequence of water injection data corresponding to each well is constructed to form the sample set fusing water injection information as model input, and the learning curve of the model is shown in Figure 9.

By comparing the prediction results of 30 wells under the two cases, it was found that the effect of fusing the water injection sequence data become both better and worse. The comparison of the prediction results of three wells is shown in Figure 10, and the corresponding evaluation indexes are shown in Table 4. The maximum lagged correlation coefficients of the adding monthly water injection series data and the corresponding monthly oil production data are 0.65689 and 0.71105 for Well 1 and Well 3, respectively, and the prediction results of both wells improved after adding the monthly water injection data. The maximum lagged correlation coefficient of the adding monthly water injection series data and the monthly oil production data is 0.34605 for well 2, and the prediction result becomes worse after adding the monthly water injection data. The experiment shows that the water injection well with the highest correlation to the target oil well found by the algorithm is not necessarily its neighboring well in practice, so there are both good and bad results after adding the water injection data.

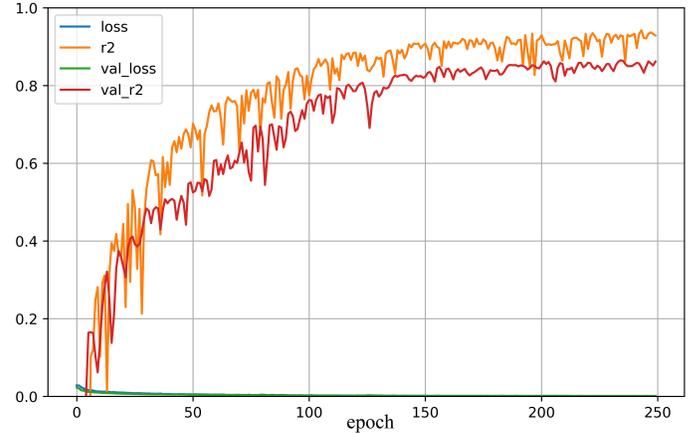

Fig. 9. Learning Curve of BP-LSTM Stacking Neural Network Trained with Samples Fusing Spatial Information

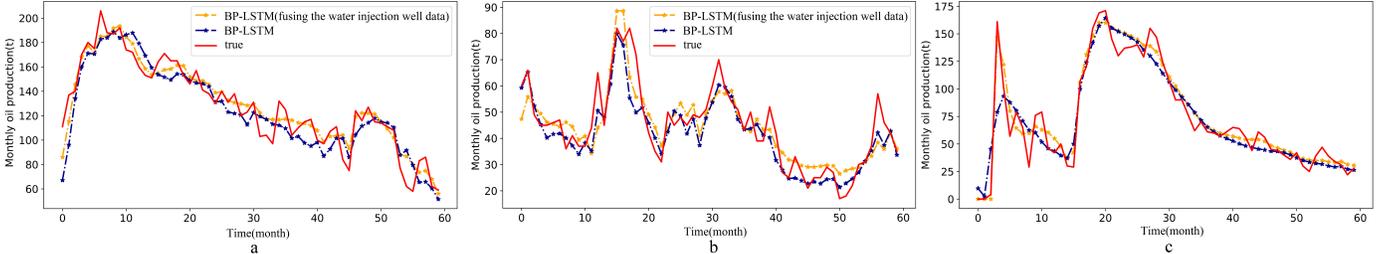

Fig. 10. Prediction Results of BP-LSTM Stacking Neural Network Trained with Samples Fusing and No Fusing Water-Injected Sequences in Well 1, Well 2 And Well 3.

TABLE 4
EVALUATION INDICATORS PREDICTED BY STACKING MODEL WITH FUSED AND NO FUSED WATER INJECTION INFORMATION

|  | Well 1 | | | Well 2 | | | Well 3 | | |
| --- | --- | --- | --- | --- | --- | --- | --- | --- | --- |
| *Correlation coefficient* | 0.65689 | | | 0.34605 | | | 0.71105 | | |
| *Lag order* | 7 | | | 4 | | | 12 | | |
| **Our Model** | R2 | MAE | RMSE | R2 | MAE | RMSE | R2 | MAE | RMSE |
| *With Fused water injection information* | 0.82842 | 11.75891 | 14.69869 | 0.80410 | 4.76333 | 6.90388 | 0.86542 | 10.81417 | 16.73249 |
| *No fused water injection information* | 0.91168 | 8.39827 | 10.54602 | 0.75011 | 5.97568 | 7.60563 | 0.947082 | 8.41420 | 10.49215 |

3. Discussion on the length of training data

Here, we further conduct experiments on the stacking model fusing water injection information to observe the influence of the length of the input data sequence on the prediction accuracy of the model. Since the added monthly water injection series are obtained by calculating the lead-lag correlation between the monthly oil production and the monthly water injection series, and the sequence correlation is closely related to the sequence length, the model effect will be different when using samples of different lengths to train the model.

We compare the prediction results of the model with time series lengths of 12 months, 30 months, and 60 months. Figure 11 exemplifies the prediction results of the model trained by using simples with different lengths of data for two wells, and Table 5 shows the evaluation indicators of the prediction results. It is obvious from the prediction results that the model prediction effect decreases as the data shortens, which is consistent with the general law of time series model for prediction. More dynamic change information of time series data can be obtained from more dynamic data.

TABLE 5
EVALUATION INDICATORS OF TWO WELLS PREDICTED WITH THE SAMPLES OF DIFFERENT DATA LENGTHS

| | Well 1 | | | Well 1 | | |
|---|---|---|---|---|---|---|
| Length(month) | R2 | MAE | RMSE | R2 | MAE | RMSE |
| *60* | 0.905285 | 2.756831 | 3.320521 | 0.881766 | 2.919449 | 4.070323 |
| *30* | 0.820716 | 3.713261 | 4.670387 | 0.739186 | 3.662061 | 5.599680 |
| *12* | 0.825806 | 5.634459 | 7.235257 | 0.613424 | 4.534502 | 5.817117 |

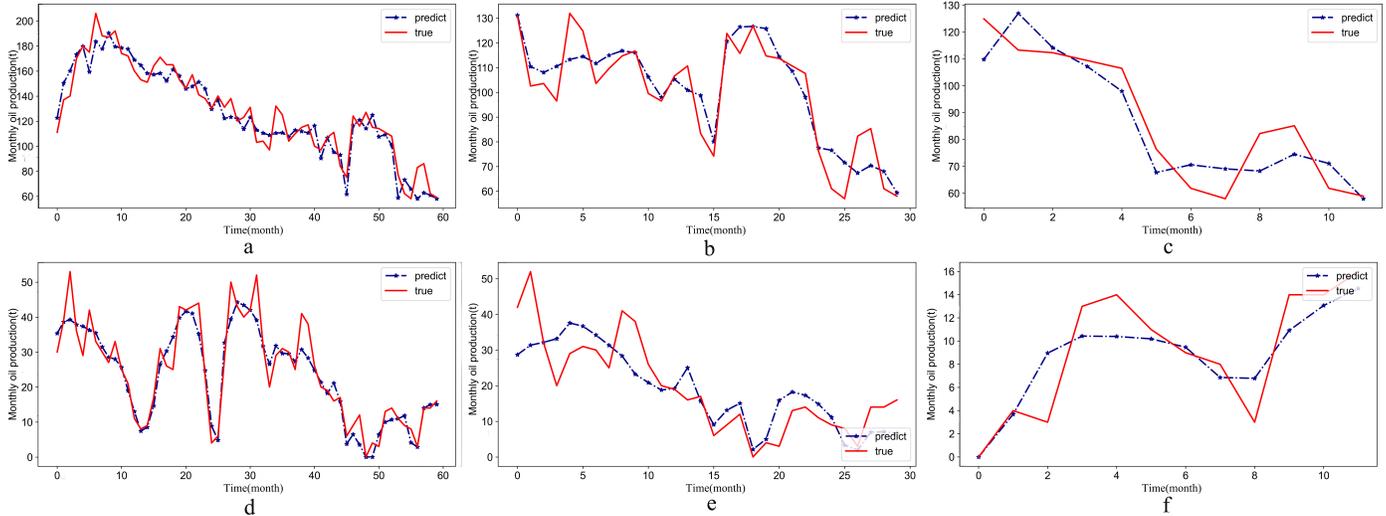

Fig. 11. Prediction Results of Two Wells Using the BP-LSTM Stacking Model Fused with Adjacent Well Water Injection Information for Training Data with Lengths of 60, 30 and 12 Months.

4. The application of the stacking model on early warning

The predicted production of an oil well using the stacking model is shown in Figure 12. We can see that there is a large deviation between the predicted and actual values starting from the 49th month onwards, with the amount of 17.6 t. By inspecting of the original data, we find that this well had taken a perforating operation at that time point. It means that in practical the practitioners really noticed the deviation of the true production to the expected production, and had to implement the stimulation treatment. This finding confirms the reliability of the presented model and can be applied to make early warning for the oil well production.

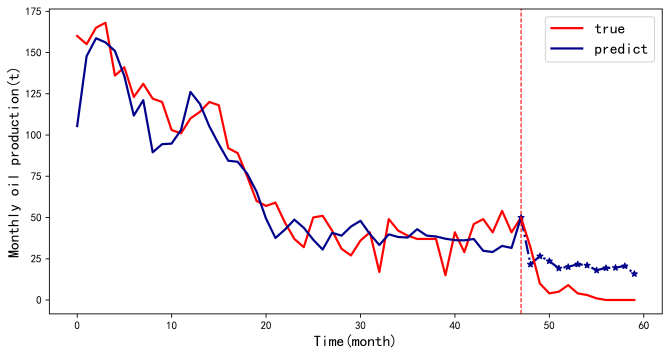

Fig. 12. Prediction Results of BP-LSTM Stacking Network

## VI. CONCLUSION

In view of the poor generalization ability of traditional machine learning predictive models, in this study we propose a novel stacking model to predict the production of the oil wells in water flooding oilfield. The main contributions of the model are as follows:

(1) The model greatly improves the prediction accuracy by fusing the static geological information, dynamic well production history, and spatial information of the adjacent water injection wells.

(2) The construction of stacking model integrates the practitioners' knowledge of the process of oil well dynamic analysis, which shows the causal link between factors and production. Based on causality discovery theory, STE is employed to quantitatively prove that the fusion of dynamic and static information has an enhanced effect on model prediction.

(3) The prediction results of stacking model can also be used for early warning of production effects to guide the optimization of the oil well development project.

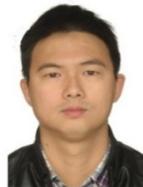

**Chao Min** was born in Xindu, Chengdu, China in 1982, Member of IEEE. He received his MSc degree in Pure Mathematics in 2007 and the Ph.D. degree in Operations and Control Theory in 2013 from Sichuan University in Chengdu, China. He is currently a Professor with the School of Sciences, Southwest Petroleum University, China. His research interests include uncertainty theory and its application, especially in petroleum engineering and data processing. He was a recipient of the Scientific and Technological Progress Award(2rd Class) of China National Association for Automation in Petroleum and Chemical Industry (CNPCI) in 2019, and the Scientific and Technological Progress Award(1st Class) of Ministry of Education of PRC in 2015.

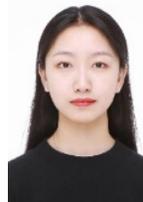

**Yijia Wang** was born in Hanzhong, Shaanxi, China, in 2000. She received the B.S. degree in Information and Computing Sciences in 2021 from Southwest Petroleum University, Chengdu, China. She is currently working towards the M.S. degree at Southwest Petroleum University, Chengdu, China. Her research interests include data analysis techniques related to oilfield development.

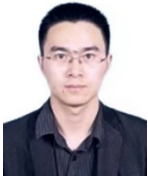
**Huohai Yang** was born in Suining, Sichuan, China in 1986. Served as the expert of the preparation of the application guide for clean energy projects of the Department of Science and Technology of Sichuan Province. He received his MSc degree in Oil-Gas Well Engineering from Southwest Petroleum University in 2012 and the Ph.D. degree in Oil-Gas Field Development Engineering in 2017 from Chengdu University of Technology in Chengdu, China. He is currently a associate professor with the school of Oil and gas engineering, Southwest Petroleum University, China. His research fields include the theory and technology research of marine oil and gas resources geological exploration and efficient development and utilization. He was a recipient of the provincial and ministerial second prize in science and technology in June 2020 (ranking second).

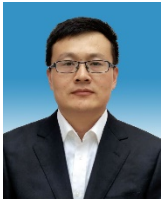
**Wei Zhao** was born in Taian, Shandong, China. He received his MSc degree in 2010 from China University of Petroleum (East China), China. He is a senior engineer and currently mainly engaged in the comprehensive strategy research of oil and gas field development. He was a recipient of Science and Technology Progress Award(2rd Class) of Sinopec, and the Scientific and Technological Progress Award(2st Class) of Shengli Oilfield.